\documentclass{article}

\usepackage{iclr2017_conference, times}

\usepackage{graphicx}
\usepackage{wrapfig}
\usepackage{caption}
\usepackage{subcaption}

\iclrfinalcopy

\usepackage{algorithm}
\usepackage{algorithmic}

\usepackage{verbatim}

\title{Exploring Sparsity in \\ Recurrent Neural Networks}

\author{Sharan Narang, Erich Elsen\thanks{Now at Google Brain eriche@google.com}, Greg Diamos \& Shubho Sengupta\thanks{Now at Facebook AI Research ssengupta@fb.com} \\
Baidu Research\\
\texttt{\{sharan,gdiamos\}@baidu.com}
}

\begin{document}

\maketitle

\begin{abstract}
Recurrent Neural Networks (RNN) are widely used to solve a variety of problems and as the quantity of data and the amount of available compute have increased, so have model sizes. The number of parameters in recent state-of-the-art networks makes them hard to deploy, especially on mobile phones and embedded devices. The challenge is due to both the size of the model and the time it takes to evaluate it.  In order to deploy these RNNs efficiently, we propose a technique to reduce the parameters of a network by pruning weights during the initial training of the network. At the end of training, the parameters of the network are sparse while accuracy is still close to the original dense neural network. The network size is reduced by 8$\times$ and the time required to train the model remains constant. Additionally, we can prune a larger dense network to achieve better than baseline performance while still reducing the total number of parameters significantly. Pruning RNNs reduces the size of the model and can also help achieve significant inference time speed-up using sparse matrix multiply. Benchmarks show that using our technique model size can be reduced by 90\% and speed-up is around 2$\times$ to 7$\times$. 
\end{abstract}

\section{Introduction}

Recent advances in multiple fields such as speech recognition~\citep{graves2014towards, deepspeech2}, language modeling~\citep{DBLP:journals/corr/JozefowiczVSSW16} and machine translation~\citep{DBLP:journals/corr/WuSCLNMKCGMKSJL16} can be at least partially attributed to larger training datasets, larger models and more compute that allows
larger models to be trained on larger datasets.

For example, the deep neural network used for acoustic modeling in~\citet{deepspeech1} had 11 million parameters which grew to approximately 67 million for bidirectional RNNs and further to 116 million for the latest forward only GRU models in~\citet{deepspeech2}. And in language modeling the size of the non-embedding parameters (mostly in the recurrent layers) have exploded even as various ways of hand engineering sparsity into the embeddings have been explored in~\citet{DBLP:journals/corr/JozefowiczVSSW16} and~\citet{DBLP:journals/corr/ChenGA15}.

These large models face two significant challenges in deployment. Mobile phones and embedded devices have limited memory and storage and in some cases network bandwidth is also a concern. In addition, the evaluation of these models requires a significant amount of computation. Even in cases when the networks can be evaluated fast enough, it will still have a significant impact on battery life in mobile devices~\citep{han2015deep}. 

Inference performance of RNNs is dominated by the memory bandwidth of the hardware, since most of the work is simply reading in the parameters at every time step. Moving from a dense calculation to a sparse one comes with a penalty, but if the sparsity factor is large enough, then the smaller amount of data required by the sparse routines becomes a win.  Furthermore, this suggests that if the parameter sizes can be reduced to fit in cache or other very fast memory, then large speedups could be realized, resulting in a super-linear increase in performance.

The more powerful server class GPUs used in data centers can generally perform inference quickly enough to serve one user, but in the data center performance per dollar is very important. Techniques that allow models to be evaluated faster enable more users to be served per GPU increasing the effective performance per dollar. 

We propose a method to reduce the number of weights in recurrent neural networks. While the  network is training we progressively set more and more weights to zero using a monotonically increasing threshold. By controlling the shape of the function that maps iteration count to threshold value, we can control how sparse the final weight matrices become. We prune all the weights of a recurrent layer; other layer types with significantly fewer parameters are not pruned. Separate threshold functions can be used for each layer, although in practice we use one threshold function per layer type. With this approach, we can achieve sparsity of ~90\% with a small loss in accuracy. We show this technique works with Gated Recurrent Units (GRU)~\citep{cho2014learning} as well as vanilla RNNs.

In addition to the benefits of less storage and faster inference, this technique can also improve the accuracy over a dense baseline. By starting with a larger dense matrix than the baseline and then pruning it down, we can achieve equal or better accuracy compared to the baseline but with a much smaller number of parameters.

This approach can be implemented easily in current training frameworks and is agnostic to the optimization algorithm. Furthermore, training time does not increase unlike previous approaches such as in~\cite{han2015deep}. State of the art results in speech recognition generally require days to weeks of training time, so a further 3-4$\times$ increase in training time is undesirable.



\section{Related Work}

There have been several proposals to reduce the memory footprint of weights and activations in neural networks. One method is to use a fixed point representation to quantize weights to signed bytes and activations to unsigned bytes~\citep{37631}. Another technique that has been tried in the past is to learn a low rank factorization of the weight matrices. One method is to carefully construct one of the factors and learn the other ~\citep{DBLP:journals/corr/DenilSDRF13}. Inspired by this technique, a low rank approximation for the convolution layers achieves twice the speed while staying within 1\% of the original model in terms of accuracy~\citep{DBLP:journals/corr/DentonZBLF14}. The convolution layer can also be approximated by a smaller set of basis filters~\citep{DBLP:journals/corr/JaderbergVZ14}. By doing this they achieve a 2.5x speedup with no loss in accuracy. Quantization techniques like k-means clustering of weights can also reduce the storage size of the models by focusing only on the fully connected layers~\citep{DBLP:journals/corr/GongLYB14}. A hash function can also reduce memory footprint by tying together weights that fall in the same hash bucket~\citep{DBLP:journals/corr/ChenWTWC15}. This reduces the model size by a factor of 8.

Yet another approach to reduce compute and network size is through network pruning. One method is to use several bias techniques to decay weights~\citep{Hanson:1989:CBM:89851.89872}. Yet another approach is to use the diagonal terms of a Hessian matrix to construct a saliency threshold and used this to drop weights that fall below a given saliency threshold~\citep{Cun90optimalbrain}. In this technique, once a weight has been set to 0, the network is retrained with these weights frozen at 0. Optimal Brain Surgeon is another work in the same vein that prunes weights using the inverse of a Hessian matrix with the additional advantage of no re-training after pruning~\citep{hassibi1993optimal}. 

Both pruning and quantization techniques can be combined to get impressive gains on AlexNet trained on the ImageNet dataset~\citep{han2015deep}. In this case, pruning, quantization and subsequent Huffman encoding results in a 35x reduction in model size without affecting accuracy. There has also been some recent work to shrink model size for recurrent and LSTM networks used in automatic speech recognition (ASR)~\citep{DBLP:journals/corr/LuSS16}. By using a hybrid strategy of using Toeplitz matrices for the bottom layer and shared low-rank factors on the top layers, they were able to reduce the parameters of a LSTM by 75\% while incurring a 0.3\% increase in word error rate (WER). 

Our method is a pruning technique that is computationally efficient for large recurrent networks that have become the norm for automatic speech recognition. Unlike the methods that need to approximate a Hessian~\citep{Cun90optimalbrain, hassibi1993optimal} our method uses a simple heuristic to choose the threshold used to drop weights. Yet another advantage, when compared to methods that need re-training~\citep{han2015deep}, is that our pruning technique is part of training and needs no additional re-training. Even though our technique requires judicious choice of pruning hyper-parameters, we feel that it is easier than choosing the structure of matrices to guide the sparsification for recurrent networks~\citep{DBLP:journals/corr/LuSS16}. Another approach for pruning feed forward neural networks for speech recognition is using simple threshold to prune all weights~\citep{yu2012exploiting} at a particular epoch. However, we find that gradual pruning produces better results than hard pruning.


\section{Implementation}

Our pruning approach involves maintaining a set of masks, a monotonically increasing threshold and a set of hyper parameters that are used to determine the threshold. During model initialization, we create a set of binary masks, one for each weight in the network that are all initially set to one. After every optimizer update step, each weight is multiplied with its corresponding mask. At regular intervals, the masks are updated by setting all parameters that are lower than the threshold to zero.

\begin{table}[t]
\caption{Hyper-Parameters used for determining threshold ($\epsilon$)}
\label{tab:hyper-param}
\begin{center}
 \begin{tabular}{p{1cm}p{5cm}p{4cm}}
\multicolumn{1}{c}{\bf HYPER-PARAM}  &\multicolumn{1}{c}{\bf DESCRIPTION} &\multicolumn{1}{c}{\bf HEURISTIC VALUES}
\\ \hline \\
 \emph{start\_itr} & Iteration to start pruning & Start of second epoch\\  
 \emph{ramp\_itr} & Iteration to increase the rate of pruning & Start of 25\% of total epochs\\ 
 \emph{end\_itr} & Iteration to stop pruning more parameters & Start of 50\% of total epochs \\ 
 \emph{start\_slope} ($\theta$) & Initial slope to prune the weights & See equation \ref{eq:start_slope} \\ 
 \emph{ramp\_slope} ($\phi$) & Ramp slope to change the rate of pruning  & $1.5 \theta$ to $2 \theta$ \\ 
 \emph{freq} & Number of iterations after which $\epsilon$ is updated & 100 \\ \hline
 \end{tabular} 
\end{center}
\end{table}

The threshold is computed using  hyper-parameters shown in Table~\ref{tab:hyper-param}. The hyper-parameters control the duration, rate and frequency of pruning the parameters for each layer. We use a different set of hyper-parameters for each layer type resulting in a different threshold for each layer type.
The threshold is updated at regular intervals using the hyper-parameters according to Algorithm~\ref{prune-algo}. We don't modify the gradients in the back-propagation step. It is possible for the updates of a pruned weight to be larger than the threshold of that layer. In this case, the weight will be involved in the forward pass again. 

We provide heuristics to help determine \emph{start\_itr}, \emph{ramp\_itr} and \emph{end\_itr} in table \ref{tab:hyper-param}. After picking these hyper parameters and assuming that \emph{ramp\_slope}($\phi$) is 1.5$\times$ \emph{start\_slope} ($\theta$), we calculate ($\theta$) using equation \ref{eq:start_slope}.

\begin{equation} \label{eq:start_slope}
    \theta = \frac{2 * q * \mathit{freq}}  {2*(\mathit{ramp\_itr} - \mathit{start\_itr}) + 3*(\mathit{end\_itr} - \mathit{ramp\_itr})}
\end{equation}

In order to determine \textit{q} in equation \ref{eq:start_slope}, we use an existing weight array from a previously trained model. The weights are sorted using absolute values and we pick the weight corresponding to the 90th percentile as \textit{q}. This allows us to pick reasonable values for the hyper-parameters required for pruning. A validation set can be used to fine tune these parameters.



We only prune the weights of the recurrent and linear layers but not the biases or batch norm parameters since they are much fewer in number compared to the weights. For the recurrent layers, we prune both the input weight matrix and the recurrent weight matrix. Similarly, we prune all the weights in gated recurrent units including those of the reset and update gates.

\begin{algorithm}[t]
  \caption{Pruning Algorithm}
  \label{prune-algo}
  \begin{algorithmic}
    \STATE $\mathit{current\_itr} = 0$ 
    \WHILE{training}
    \FORALL {parameters}
    \STATE $\mathit{param} =(\mathit{param}$ \AND $\mathit{mask}$)
      \IF{$\mathit{current\_itr} > \mathit{start\_itr}$ \AND $\mathit{current\_itr} < \mathit{end\_itr}$}
      \IF{$(\mathit{current\_itr} \bmod \mathit{freq})$ == 0}
      \IF{$\mathit{current\_itr} < \mathit{ramp\_itr}$}
        \STATE $\epsilon = \theta * (\mathit{current\_itr} - \mathit{start\_itr} + 1 )/ \mathit{freq}$
      \ELSE
        \STATE $\epsilon = (\theta * (\mathit{ramp\_itr} - \mathit{start\_itr} + 1) + \phi * (\mathit{current\_itr} - \mathit{ramp\_itr} + 1))/\mathit{freq} $
      \ENDIF
      \STATE $\mathit{mask} = \mathit{abs}(\mathit{param}) < \epsilon$
      \ENDIF
      \ENDIF
      \ENDFOR
      \STATE $\mathit{current\_itr} \mathrel{+}=1$
    \ENDWHILE
  \end{algorithmic}
\end{algorithm}

\section{Experiments}
We run all our experiments on a training set of 2100 hours of English speech data and a validation set of 3.5 hours of multi-speaker data. This is a small subset of the datasets that we use to train our state-of-the-art automatic speech recognition models. We train the models using Nesterov SGD for 20 epochs. Besides the hyper-parameters for determining the threshold, all other hyper-parameters remain unchanged between the dense and sparse training runs. We find that our pruning approach works well for vanilla bidirectional recurrent layers and forward only gated recurrent units.  

\subsection{Bidirectional RNNs}
We use the Deep Speech 2 model for these experiments. As shown in Table~\ref{ds2-arch}, this model has 2 convolution layers, followed by 7 bidirectional recurrent layers and a CTC cost layer. Each recurrent linear layer has 1760 hidden units, creating a network of approximately 67 million parameters. For these experiments, we prune the linear layers that feed into the recurrent layers, the forward and backward recurrent layers and fully connected layer before the CTC layer. These experiments use clipped rectified-linear units (ReLU) $\sigma(x) = \min ( \max (x, 0) ,20)$ as the activation function. 

In the sparse run, the pruning begins shortly after the first epoch and continues until the $\textrm{10}^{\textrm{th}}$ epoch. We chose these hyper-parameters so that the model has an overall sparsity of 88\% at the end of pruning, which is 8x smaller than the original dense model. The character error rate (CER) on the devset is about 20\% worse relative to the dense model as shown in Table \ref{tab:results}. 

An argument against this sparsity result might be that we are taking advantage of a large model that overfits our relatively small dataset. In order to test this hypothesis, we train a dense model with 704 hidden units in each layer, that has approximately the same number of parameters as the final sparse model. Table~\ref{tab:results} shows that this model performs worse than the sparse models. Thus sparse model is a better approach to reduce parameters than using a dense model with fewer hidden units. 

\begin{figure}[t]
 
\begin{subfigure}{0.49\linewidth}
\includegraphics[width=0.9\linewidth]{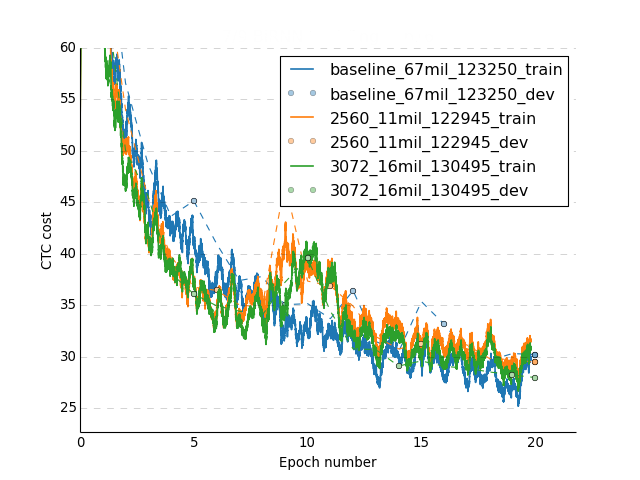} 
\caption{}
\label{fig:rnn}
\end{subfigure}
\begin{subfigure}{0.49\linewidth}
\includegraphics[width=0.9\linewidth]{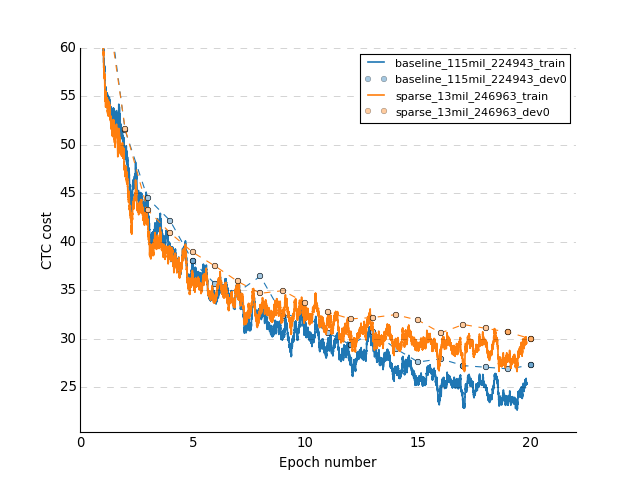}
\caption{}
\label{fig:gru-results}
\end{subfigure}
 
\caption{Training and dev curves for baseline (dense) and sparse training. Figure \ref{fig:rnn} includes training and dev curves for models with larger recurrent layers with 2560 and 3072 hidden units compared to the 1760 dense baseline. Figure \ref{fig:gru-results} plots the training and dev curves for GRU models (sparse and dense) with 2560 parameters.}
\label{fig:results}
\end{figure}

In order to recover the loss in accuracy, we train sparse models with larger recurrent layers with 2560 and 3072 hidden units. Figure~\ref{fig:rnn} shows the training and dev curves for these sparse models compared to the dense baseline model. These experiments use the same hyper-parameters (except for small changes in the pruning hyper-parameters) and the same dataset as the baseline model. As we see in Table~\ref{tab:results}, the model with 2560 hidden units achieves a 0.75\% relative improvement compared to the dense baseline model, while the model with 3072 hidden units has a 3.95\% improvement. The dense 2560 model also improves the CER by 11.85\% relative to the dense baseline model. The sparse 2560 model is about 12\% worse than the corresponding dense model. Both these large models are pruned to achieve a final sparsity of around 92\%. These sparse larger models have significantly fewer parameters than the baseline dense model. 

We also compare our gradual pruning approach to the hard pruning approach proposed in \citet{yu2012exploiting}. In their approach, all parameters below a certain threshold are pruned at particular epoch. Table \ref{tab:hard_thresh_results} shows the results of pruning the RNN dense baseline model at different epochs to achieve final parameter count ranging from 8 million to 11 million. The network is trained for the same number of epochs as the gradual pruning experiments. These hard threshold results are compared with the RNN Sparse 1760 model in Table \ref{tab:results}. For approximately same number of parameters, gradual pruning is 7\% to 9\% better than hard pruning.

We conclude that pruning models to achieve sparsity of around 90\% reduces the relative accuracy of the model by 10\% to 20\%. However, for a given performance requirement, it is better to prune a larger model than to use a smaller dense model. Gradually pruning a model produces better results than hard pruning. 

\begin{table}[t]
\caption{Deep Speech 2 architecture with 1760 hidden units}
\label{ds2-arch}
\begin{center}
 \begin{tabular}{llll}
\multicolumn{1}{c}{\bf LAYER ID}  &\multicolumn{1}{c}{\bf TYPE} &\multicolumn{1}{c}{\bf \# PARAMS}
\\ \hline \\
layer 0  & 2D Convolution          &    19616\\
layer 1  & 2D Convolution          &   239168\\
layer 2  & Bidirectional Recurrent Linear &  8507840\\
layer 3  & Bidirectional Recurrent Linear &  9296320\\
layer 4  & Bidirectional Recurrent Linear &  9296320\\ 
layer 5  & Bidirectional Recurrent Linear &  9296320\\
layer 6  & Bidirectional Recurrent Linear &  9296320\\
layer 7  & Bidirectional Recurrent Linear &  9296320\\ 
layer 8  & Bidirectional Recurrent Linear &  9296320\\ 
layer 9  & FullyConnected          &  3101120\\
layer 10  & CTCCost     &   95054 \\ \hline
\end{tabular}
\end{center}
\end{table}

\begin{table}[h]
\caption{GRU \& bidirectional RNN model results}
\label{tab:results}
\begin{center}
 \begin{tabular}{lllll}
\multicolumn{1}{c}{\bf MODEL} & \multicolumn{1}{c}{\bf \# UNITS}  &\multicolumn{1}{c}{\bf CER} &\multicolumn{1}{c}{\bf \# PARAMS} &\multicolumn{1}{c}{\bf RELATIVE PERF} 
\\ \hline \\
RNN Dense Baseline & 1760 & 10.67 & 67 million & 0.0\% \\
RNN Dense Small & 704 & 14.50 & 11.6 million & -35.89\% \\
RNN Dense Medium & 2560 & 9.43 & 141 million & 11.85\% \\
RNN Sparse 1760 & 1760 & 12.88 & 8.3 million & -20.71\% \\
RNN Sparse Medium & 2560 & \bf{10.59} & 11.1 million & 0.75\% \\
RNN Sparse Big & 3072 & \bf{10.25} & 16.7 million & 3.95\% \\ 
\hline
GRU Dense & 2560 & 9.55 & 115 million & 0.0\% \\
GRU Sparse & 2560 & 10.87 & 13 million & -13.82\% \\
GRU Sparse Medium & 3568 & \bf{9.76} & 17.8 million & -2.20\%\\
\hline
\end{tabular}
\end{center}
\end{table}

\begin{table}[h]
\caption{RNN dense baseline model with hard pruning}
\label{tab:hard_thresh_results}
\begin{center}
 \begin{tabular}{lllll}
\multicolumn{1}{c}{\bf \# UNITS} 
&\multicolumn{1}{c}{\bf PRUNED EPOCH}
&\multicolumn{1}{c}{\bf CER} &\multicolumn{1}{c}{\bf \# PARAMS} &\multicolumn{1}{c}{\bf RELATIVE PERF} 
\\ \hline \\
1760 & 5 & 13.82 & 8 million & -29.52\% \\
1760 & 7 & 13.27 & 11 million & -24.37\% \\
1760 & 10 & 13.41 & 8.4 million & -25.68\% \\
1760 & 12 & 13.63 & 8 million & -27.74\% \\
1760 & 15 & 26.33 & 9.2 million & -146.77\% \\
\hline
\end{tabular}
\end{center}
\end{table}


\subsection{Gated Recurrent Units}

We also experimented with GRU models shown in Table~\ref{gru-arch}, that have 2560 hidden units in the GRU layer and a total of 115 million parameters. For these experiments, we prune all layers except the convolution layers since they have relatively fewer parameters. 

Figure~\ref{fig:gru-results} compares the training and dev curves of a sparse GRU model a dense GRU model. The sparse GRU model has a 13.8\% drop in the accuracy relative to the dense model. As shown in Table~\ref{tab:results}, the sparse model has an overall sparsity of 88.6\% with 13 million parameters. Similar to the RNN models, we train a sparse GRU model with 3568 hidden units. The dataset and the hyperparameters are not changed from the previous GRU experiments. This model has an overall sparsity of 91.82\% with 17.8 million parameters. As shown in Table \ref{tab:results}, the model with 3568 hidden units is only 2.2\% worse than the baseline dense GRU model. We expect to match the performance of the GRU dense network by slightly lowering the sparsity of this network or by increasing the hidden units for the layers. 

In addition, we experimented with pruning only the GRU layers and keeping all the parameters in fully connected layers. The accuracy for these experiments is around 7\% worse than the baseline dense model. However, this model only achieves 50\% compression due to the size of the fully connected layers.

\begin{table}[t]
\caption{Gated recurrent units model}
\label{gru-arch}
\begin{center}
 \begin{tabular}{lll}
\multicolumn{1}{c}{\bf LAYER ID}  &\multicolumn{1}{c}{\bf TYPE} &\multicolumn{1}{c}{\bf \# PARAMS}
\\ \hline \\
layer 0  & 2D Convolution          &    19616\\
layer 1  & 2D Convolution          &   239168\\
layer 2  & Gated Recurrent Linear &  29752320\\
layer 3  & Gated Recurrent Linear &  39336960\\
layer 4  & Gated Recurrent Linear &  39336960\\ 
layer 5  & Row Convolution & 107520 \\
layer 6  & FullyConnected          &  6558720\\
layer 7  & CTCCost     &   74269 \\ \hline
\end{tabular}
\end{center}
\end{table}



\section{Performance}
\subsection{Compute time}
The success of deep learning in recent years have been driven by large models trained on large datasets. However this also increases the inference time after the models have been deployed. We can mitigate this effect by using sparse layers. 

A General Matrix-Matrix Multiply (GEMM) is the most compute intensive operation in evaluating a neural network model. Table~\ref{gemm-time} compares times for GEMM for recurrent layers with different number of hidden units that are 95\% sparse. The performance benchmark was run using NVIDIA's CUDNN and cuSPARSE libraries on a TitanX Maxwell GPU and compiled using CUDA 7.5. All experiments are run on a minibatch of 1 and in this case, the operation is known as a sparse matrix-vector product (SpMV). We can achieve speed-ups ranging from 3x to 1.15x depending on the size of the recurrent layer. Similarly, for the GRU models, the speed-ups range from 7x to 3.5x. However, we notice that cuSPARSE performance is substantially lower than the approximately 20x speedup that we would expect by comparing the bandwidth requirements of the 95\% sparse and dense networks. State of the art SpMV routines can achieve close to device memory bandwidth for a wide array of matrix shapes and sparsity patterns (see~\cite{ModernGPU} and~\cite{Liu:2013:ESM:2464996.2465013}).  This means that the performance should improve by the factor that parameter counts are reduced. Additionally, we find that the cuSPARSE performance degrades with larger batch sizes. It should be possible for a better implementation to further exploit the significant reuse of the weight matrix provided by large batch sizes.   
\begin{table}[t]
\caption{GEMM times for recurrent layers with different sparsity}
\label{gemm-time}
\begin{center}
\begin{tabular}{lllll}
\multicolumn{1}{c}{\bf LAYER SIZE}  &\multicolumn{1}{c}{\bf SPARSITY} &\multicolumn{1}{c}{\bf LAYER TYPE} &\multicolumn{1}{c}{\bf TIME ($\mu$sec)} &\multicolumn{1}{c}{\bf SPEEDUP}
\\ \hline \\
1760 & 0\% & RNN & 56 & 1\\
1760 & 95\% & RNN & 20 & 2.8\\
2560 & 95\% & RNN & 29 & 1.93\\
3072 & 95\% & RNN & 48 & 1.16\\ \hline
2560 & 0\% & GRU & 313 & 1\\
2560 & 95\% & GRU & 46 & 6.80\\
3568 & 95\% & GRU & 89 & 3.5\\
\hline
\end{tabular}
\end{center}
\end{table}

\subsection{Compression}
Pruning allows us to reduce the memory footprint of a model which allows them to be deployed on phones and other embedded devices. The Deep Speech 2 model can be compressed from 268 MB to around 32 MB (1760 hidden units) or 64 MB (3072 hidden units). The GRU model can be compressed from 460 MB to 50 MB. These pruned models can be further quantized down to float16 or other smaller datatypes to further reduce the memory requirements without impacting accuracy.

\begin{figure}[t]
\begin{subfigure}{0.49\linewidth}
\includegraphics[width=0.9\linewidth]{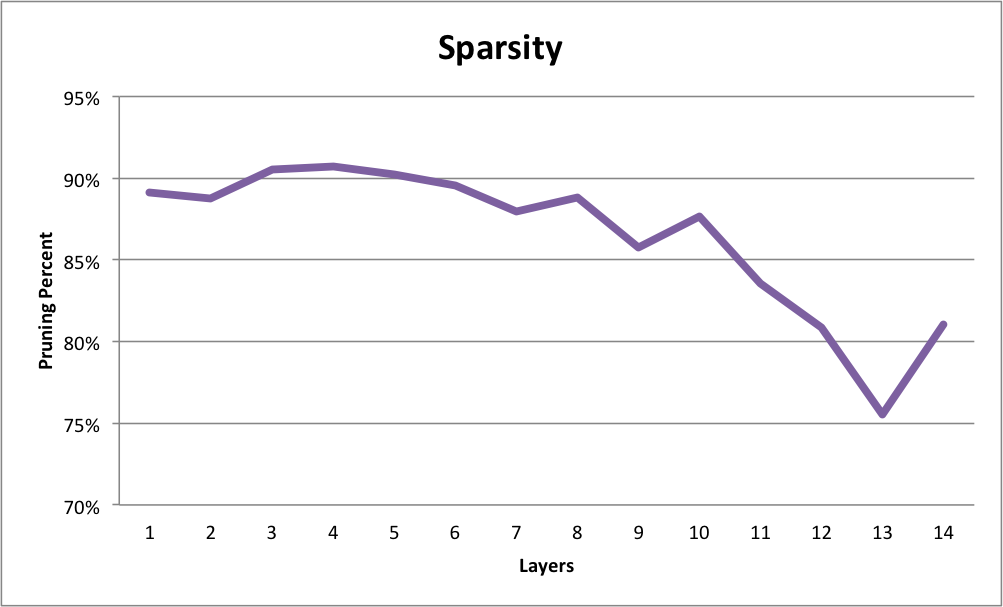} 
\caption{}
\label{fig:percent_layers}
\end{subfigure}
\begin{subfigure}{0.49\linewidth}
\includegraphics[width=0.9\linewidth]{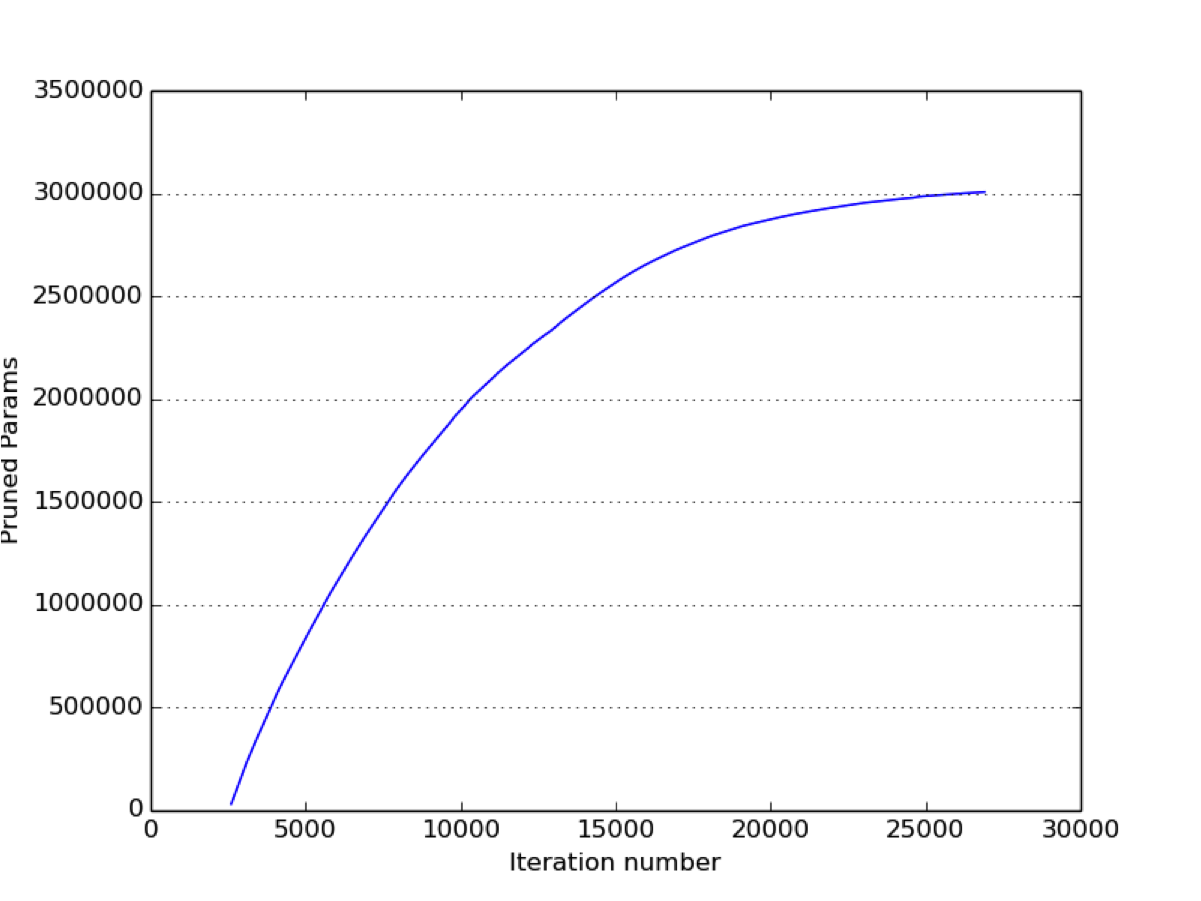}
\caption{}
\label{fig:pruning_sch}
\end{subfigure}
\caption{Pruning characteristics. Figure \ref{fig:percent_layers} plots sparsity of recurrent layers in the network with the same hyper-parameters used for pruning
. Figure~\ref{fig:pruning_sch} plots the pruning schedule of a single layer during a training run. }
\label{fig:discussion}
\end{figure}


\section{Discussion}
\subsection{Pruning Characteristics}
\label{sec:prune-char}
Figure~\ref{fig:percent_layers} shows the sparsity of all the recurrent layers with the same hyper-parameters used to prune the layers. The layers are ordered such that layer 1 is closest to input and layer 14 is the final recurrent layer before the cost layer. We see that the initial layers are pruned more aggressively compared to the final layers. We also performed experiments where the hyper parameters are different for the recurrent layers resulting in equal sparsity for all the layers. However, we get higher CER for these experiments. We conclude that to get good accuracy, it is important to prune the final layers slightly less than the initial ones. 

In Figure~\ref{fig:pruning_sch}, we plot the pruning schedule of a 95\% sparse recurrent layer of the bidirectional model trained for 20 epochs (55000 iterations). We begin pruning the network at the start of the second epoch at 2700 iterations. We stop pruning a layer after 10 epochs (half the total epochs) are complete at 27000 iterations. We see that nearly 25000 weights are pruned before 5 epochs are complete at around 15000 iterations. In our experiments, we've noticed that pruning schedules that are a convex curve tend to outperform schedules with a linear slope. 

\subsection{Persistent Kernels}
Persistent Recurrent Neural Networks~\citep{diamos2016persistent} is a technique that increases the computational intensity of evaluating an RNN by caching the weights in on-chip memory such as caches, block RAM, or register files across multiple timesteps. A high degree of sparsity allows significantly large Persistent RNNs to be stored in on-chip memory. When all the weights are stored in float16, a NVIDIA P100 GPU can support a vanilla RNN size of about 2600 hidden units. With the same datatype, at $90\%$ sparsity, and $99\%$ sparsity, a P100 can support RNNs with about 8000, and 24000 hidden units respectively. We expect these kernels to be bandwidth limited out of the memory that is used to store the parameters. This offers the potential of a 146x speedup compared to the TitanX GPU if the entire RNN layer can be stored in registers rather than the GPU DRAM of a TitanX.

Additionally, sparse matrix multiplication involves scheduling and load balancing phases to divide the work up evenly over thousands of threads and to route corresponding weights and activations to individual threads. Since the sparsity patterns for RNNs are fixed over many timesteps these scheduling and load balancing operations can be factored outside of the loop, performed once, and reused many times.

\section{Conclusion and Future Work}

We have demonstrated that by pruning the weights of RNNs during training we can find sparse models that are more accurate than dense models while significantly reducing model size. These sparse models are especially suited for deployment on mobile devices and on back-end server farms due to their small size and increased computational efficiency. Even with existing sub-optimal sparse matrix-vector libraries we realize speed-ups with these models. This technique is orthogonal to quantization techniques which would allow for even further reductions in model size and corresponding increase in performance.

We wish to investigate whether these techniques can generalize to language modeling tasks and if they can effectively reduce the size of embedding layers. We also wish to compare the sparsity generated by our pruning technique to that obtained by L1 regularization.

We are investigating training techniques that don't require maintaining dense matrices for a significant portion of the calculation. Further work remains to implement optimal small batch sparse matrix-dense vector routine for GPUs and ARM processors that would help in deployment.

\subsubsection*{Acknowledgments}
We would like to thank Bryan Catanzaro for helpful discussions related to this work. 

\bibliography{references}
\bibliographystyle{iclr2017_conference}

\end{document}